\documentclass{article}

\usepackage[dblblindworkshop,final]{neurips_2026}  
\workshoptitle{Can We Trust the Judge? Building Reliable Evaluation
for Language Models}

\usepackage[utf8]{inputenc}
\usepackage[T1]{fontenc}
\usepackage{booktabs}
\usepackage{microtype}
\usepackage{amsmath,amssymb,amsthm}
\usepackage{graphicx}
\usepackage{enumitem}
\usepackage{tabularx}
\newcommand{\ind}{\mathbf{1}}
\usepackage{url}
\usepackage{placeins}
\usepackage[hidelinks]{hyperref}

\newcommand{\E}{\mathbb{E}}
\newcommand{\Var}{\mathrm{Var}}
\newcommand{\Cov}{\mathrm{Cov}}
\newcommand{\RAMSD}{\textsc{RA-MSD}}
\newcommand{\ELDGR}{\textsc{EL-DGR}}
\newcommand{\Rlogic}{R_{\mathrm{logic}}}
\newcommand{\Rfact}{R_{\mathrm{fact}}}
\newcommand{\Repist}{R_{\mathrm{epist}}}
\newcommand{\Rinfo}{R_{\mathrm{info}}}
\newcommand{\Rtriang}{R_{\mathrm{triang}}}
\newcommand{\Radv}{R_{\mathrm{adv}}}
\newcommand{\Rmin}{R_{\mathrm{min}}}
\newcommand{\Rstep}{R_{\mathrm{step}}}

\title{When the Judge Should Not Decide:\\
Evidence-Locked, Non-Compensatory Selection\\
Bounds LLM-Judge Failure in Reasoning Pipelines}

\author{
Yiyao Zhang$^{a}$,
Diksha Goel$^{b}$,
Hussain Ahmad$^{a}$\\[0.8em]
\begin{minipage}{0.92\textwidth}
\centering\small
$^{a}$School of Computer Science and Information Technology, Adelaide University, Adelaide, SA, Australia\\
$^{b}$CSIRO's Data61, Clayton, VIC, Australia\\[0.4em]
\textit{Author e-mail addresses:}
yiyao.zhang@student.adelaide.edu.au (Y. Zhang),
diksha.goel@csiro.au (D. Goel),
hussain.ahmad@adelaide.edu.au (H. Ahmad)
\end{minipage}
}

\begin{document}
\maketitle

\begin{abstract}
An LLM judge deployed inside a reasoning pipeline does not merely
measure quality ,  it decides which answer ships. We show that the
cost of that decision depends less on judge accuracy than on the
\emph{decision rule} the judge is embedded in. On frozen candidate
pools from four GRPO policies, an unconstrained scalar
DeepSeek-R1-7B judge buys almost nothing over answer-level majority
vote ($+1.0$ pp on 500 GSM8K questions, $+0.34$ EM on 300 HotpotQA
questions), and on a frozen-rule 30-question confirmation split it is
\emph{10 points worse} than majority ,  a judge that destroys accuracy
while scoring candidates confidently. We then subordinate the same
judge to Evidence-Locked Derive--Gate--Repair (\ELDGR{}), a
task-adaptive non-compensatory rule under which a judge preference may
override evidence-supported consensus only with an extractive evidence
certificate, and a repair only when neither alternative is certified
and the repair is. With no change to the judge, the candidates, or the
budget, \ELDGR{} reaches 58.2\% on GSM8K (vs.\ 56.8\% judge, 55.8\%
majority, 55.4\% first candidate) and 17.33 EM / 25.46 F1 on HotpotQA
(vs.\ 15.67/23.49, 15.33/23.19, 15.33/22.97), improving on
first-candidate GRPO by $+2.8$ pp (exact McNemar $p{=}0.0026$) and
$+2.00$ EM ($p{=}0.070$, borderline). A decision audit shows why:
\ELDGR{} overturns consensus on only 8 of 30 pilot questions and never
converts a correct consensus into an incorrect answer. We also report
what did \emph{not} work: the same seven-channel decomposition used as
a step-level gated training reward is null, and corrected channel-drop
ablations show no channel is individually necessary ($p{=}1.0$
throughout). The practitioner-facing finding is negative about judges
and positive about admissibility ,  bound the judge's blast radius
rather than trying to make it accurate.
\end{abstract}

\section{Introduction}
\label{sec:intro}

LLM judges are increasingly load-bearing. They gate which of $G$
sampled candidates is returned, filter preference data, and supply
reward for RL post-training \citep{zheng2023judging,ouyang2022training,
shao2024deepseekmath,lightman2023let}. In each case the judge's output
is not a measurement that a human later reads ,  it is a decision that
the pipeline executes. The same is true wherever an LLM score is wired
into an operational loop: multi-LLM scoring and selection for portfolio
construction \citep{chen2025trader}, agentic triage and remediation of
software vulnerabilities \citep{arifin2026agenticvm}, and clinical
question answering, where model-to-model disagreement is large enough
that the aggregation rule is itself a safety-relevant design choice
\citep{santhosh2026healthcare}. Evaluation validity is therefore not a property
of the judge alone but of the judge \emph{plus} the rule that consumes
its score. A judge with a 60\% pairwise agreement rate is harmless if
it can only break ties among already-admissible answers, and dangerous
if it can unilaterally overwrite a correct consensus.

Most work on judge reliability targets the first half of that pair , 
measuring and correcting the judge itself, through position-bias audits
and swap averaging \citep{wang2024fair}, self-preference detection
\citep{panickssery2024selfpref}, and the calibration and agreement
protocols surveyed by \citet{gu2024judgesurvey}. We study the second
half: holding the judge, candidate pool, prompts, and compute budget
fixed and varying only the decision rule isolates a quantity judge
benchmarks do not report ,  how much of a judge's damage comes from
its scores, and how much from the authority it is granted.

\paragraph{The conflation a scalar judge cannot see.}
Consider \emph{``What is the capital of the country that hosted the
2022 FIFA World Cup?''} and the chain: (1) \emph{The 2022 FIFA World
Cup was hosted by Qatar.} (2) \emph{The capital of Qatar is Dubai.}
(3) \emph{Therefore the answer is Dubai.} Step 3 follows validly from
Step 2 and the chain reads fluently, but Step 2 is contradicted by the
first retrievable document. A judge emitting one scalar for
``quality'' scores this chain highly, because logical structure and
evidential support have been collapsed into a number that cannot
represent their disagreement ,  and the failure is silent, since the
score carries no indication of \emph{which} dimension produced it.

\paragraph{Two responses, one of which fails.}
The obvious fix is to decompose: score the step on several explicit
dimensions and recombine. We implemented this as \RAMSD{}
(Retrieval-Augmented Multi-Signal Decision reward): seven step-level
channels with logic and factuality gating the weighted auxiliary score
(Section~\ref{sec:method}). As a \emph{training} signal this did not
work ,  held-out accuracy was indistinguishable from plain GRPO, and
channel-drop ablations show no channel is individually necessary
(Section~\ref{sec:negative}). We report it because it localizes the
problem: any scheme that eventually recombines the channels into one
scalar reinherits the conflation it was built to remove.

The response that does work applies the decomposition elsewhere in the
pipeline: not as a summand, but as an \emph{admissibility constraint}
on the judge's authority. Evidence-Locked Derive--Gate--Repair
(\ELDGR{}) partitions candidates into feasibility strata by a cheap
certificate ,  an extractive occurrence check against retrieved
evidence for retrieval QA, an independently derived and re-checked
numeric answer for arithmetic. The judge's scalar preference then ranks
candidates \emph{within} a stratum but can never promote one across
strata, so an evidence-supported consensus answer cannot be displaced
by a higher-scoring but uncertified alternative.

\paragraph{Contributions.}
\begin{enumerate}[leftmargin=1.4em,itemsep=0pt,topsep=1pt]
\item \textbf{A measurement of judge authority, not judge accuracy:}
on matched frozen pools, an unconstrained scalar judge is worth $+1.0$
pp (GSM8K, $n{=}500$) and $+0.34$ EM (HotpotQA, $n{=}300$) over
majority vote, and $-10.0$ pp on a frozen-rule confirmation split
($n{=}30$), below even first-candidate selection
(Section~\ref{sec:results}).
\item \textbf{\ELDGR{}, a non-compensatory decision rule} that
subordinates the same judge to task-adaptive evidence certificates,
improving every protocol-matched selector on both datasets with no
extra candidates and $\leq 5$s/question of verifier latency
(Sections~\ref{sec:method}--\ref{sec:results}).
\item \textbf{A decision audit and a negative result:} where
\ELDGR{}'s gain comes from (8/30 consensus overrides, 0
correct-to-incorrect flips) and where the same decomposition fails as a
training reward (Sections~\ref{sec:audit}--\ref{sec:negative}).
\end{enumerate}

This is a post-training \emph{selection} study: we do not claim
\ELDGR{} trains a better policy, and absolute scores are not comparable
to published state of the art, since the candidate artifacts come from
a 3B policy under a constrained budget. All selectors consume the
identical frozen pool, so the paired contrasts are internally valid.

\section{Related Work}
\label{sec:related}

\paragraph{Judge reliability.}
The dominant framing treats evaluator quality as a measurement problem:
how well does the judge agree with humans, and which biases distort
that agreement? This line establishes pairwise-agreement benchmarks
\citep{zheng2023judging}, documents position and ordering effects with
swap-averaging remedies \citep{wang2024fair}, and shows that evaluators
prefer their own generations \citep{panickssery2024selfpref};
\citet{gu2024judgesurvey} survey the resulting protocols. We take the
judge's error profile as given ,  neither debiasing nor fine-tuning it
,  and ask what the surrounding decision rule does to the
\emph{consequences} of those errors. That second factor can dominate
the first: below, identical scores swing from 10 points below majority
vote to the best selector tested, purely through admissibility.
Constraining authority rather than improving the scorer is the same
move made by compositional shielding, where a verifier decides which
agent actions are admissible and the system abstains when none is
certified \citep{zhang2026vacs}, and by uncertainty-gated meta-reasoning,
where a competence test decides whether an auxiliary call is consulted
at all \citep{zhang2026meta}. \ELDGR{} applies that discipline to the
narrower object of a judge's selection authority over a frozen pool.

\paragraph{Step supervision and grounded verification.}
Process reward models supply dense local feedback but usually collapse
logical validity, factual support, and redundancy into one scalar
\citep{lightman2023let,wang2025math}. Retrieval-based systems use
external evidence during \emph{generation}
\citep{lewis2020retrieval,yao2023react,asai2024selfrag,nakano2021webgpt},
whereas we use retrieval at scoring time. The grounding channel is
closest to claim-verification metrics such as FActScore
\citep{min2023factscore,chern2023factool}, moved from offline
evaluation into an online decision rule; self-consistency
\citep{wang2023selfconsistency} supplies our judge-free majority
baseline.

\paragraph{Positioning against recent multi-signal methods.}
Table~\ref{tab:recent_positioning} states the comparison boundary
against five 2025 systems that also decompose or densify reasoning
feedback (Table~\ref{tab:recent_positioning},
Appendix~\ref{app:positioning}): DRM \citep{wang2025drm}, BCRL
\citep{wu2025bcrl}, GAR \citep{liu2025gar}, AutoDSPy
\citep{azim2025autodspy}, and SWiRL \citep{lu2025swirl}. Their
published scores are not copied into Table~\ref{tab:main_results}:
model, retrieval corpus, budget, and protocol all differ. Our matched
baselines are therefore \emph{selectors} that consume exactly the same
frozen candidate pool ,  the only class for which a paired test on a
fixed pool is meaningful.

\section{Setting: The Judge as a Selector}
\label{sec:setting}

\paragraph{Judge conflation.}
Let $\mathcal{D}_t$ be the documents retrieved for a reasoning step
$s_t$ and let $f_{\mathrm{NLI}}(s_t \mid d) \in [-1,1]$ map
contradiction to $-1$, neutral to $0$, and entailment to $+1$. A
step-level score $R$ suffers \emph{$(\eta,\delta)$-conflation}, for
$\delta\in(0,1]$ and $\eta<0$, if some reachable step satisfies
\begin{equation}
R(s_t) \geq \delta
\quad\text{and}\quad
\max_{d \in \mathcal{D}_t} f_{\mathrm{NLI}}(s_t \mid d) \leq \eta .
\label{eq:conflation}
\end{equation}
The Qatar/Dubai step is a severe
$(\eta{=}{-}0.8,\delta{=}0.85)$-conflation. Closed-book judges admit
conflation by construction: nothing in their input distinguishes a
supported step from a fluent unsupported one. Outcome reward models
admit it at every intermediate step; rule-based format-plus-match
rewards admit it whenever surface form and evidence support
decorrelate.

\paragraph{Selection protocol.}
Fix a question $Q$, a frozen candidate set $\{y_1,\dots,y_G\}$ with
normalized answers $a_j$, and retrieved evidence $E$. Let $m$ be the
answer-level majority candidate \citep{wang2023selfconsistency}, $s$
the candidate a scalar judge prefers, and $r$ an independently derived
and checked repair. A \emph{selector} maps $(Q,\{y_j\},E)$ to a
returned answer. Our baselines are first candidate (what a
single-sample pipeline returns), majority vote (judge-free consensus),
and the scalar judge $s$ (the judge given full authority); \ELDGR{} is
a fourth selector using the same $s$ and $E$. Because every selector is
fed the identical pool, differences measure decision rules, not
generation.

\section{Method}
\label{sec:method}

\subsection{Seven diagnostic channels}
\label{sec:channels}

\begin{figure}[t]
\centering
\includegraphics[width=0.88\textwidth]{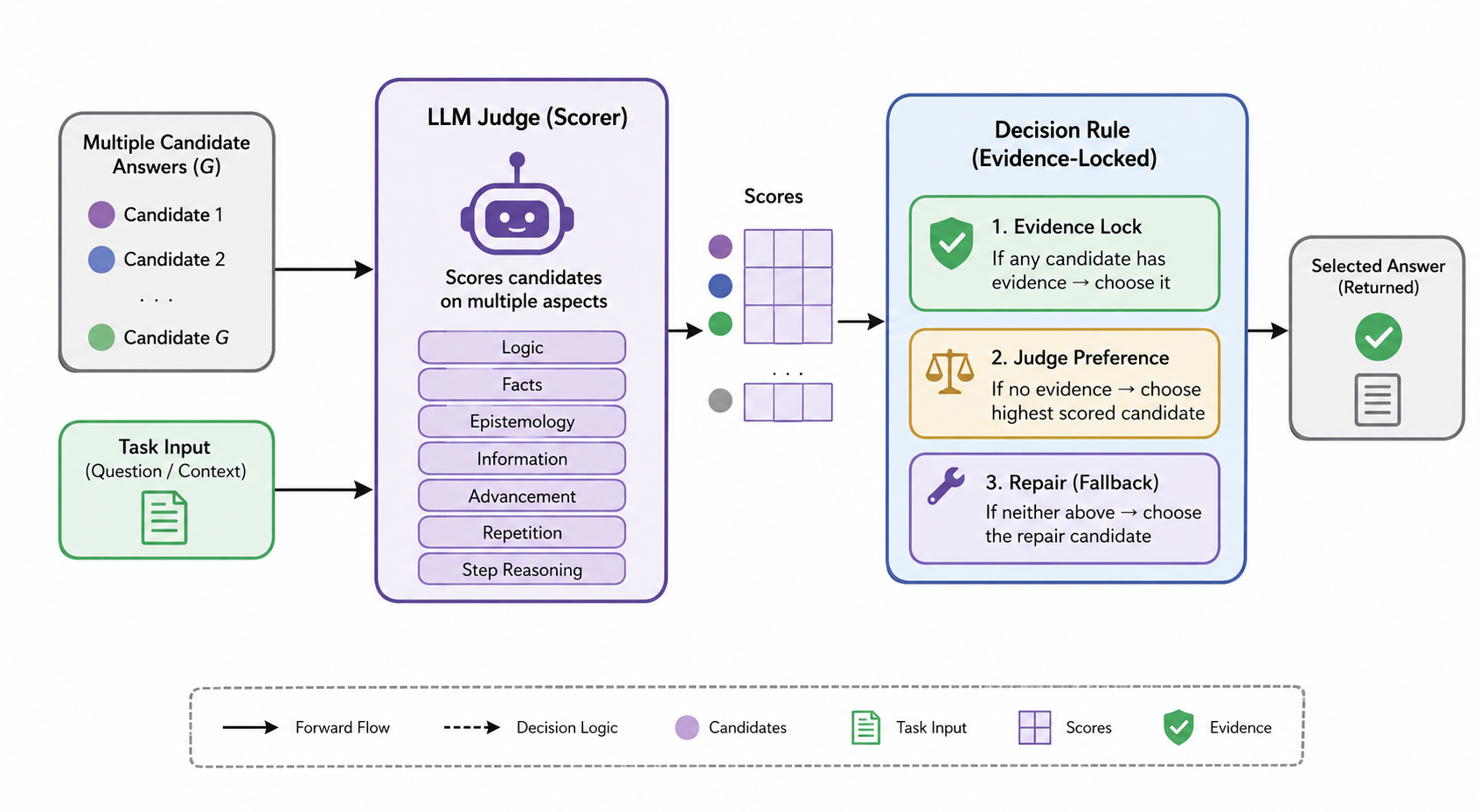}
\caption{The judge-as-selector pipeline. Candidates and task input are
scored by an LLM judge on seven diagnostic aspects
(Table~\ref{tab:signal_summary}) instead of a single scalar; an
evidence-locked rule then partitions candidates into feasibility
strata, the judge's preference ranks candidates only within a stratum,
and a checked repair is admitted only when no certified candidate
exists. The exact ordering is Eq.~\eqref{eq:eldgr}.}
\label{fig:pipeline}
\end{figure}

Figure~\ref{fig:pipeline} shows the pipeline: a frozen candidate group
and the task input enter a judge that scores each candidate on several
explicit aspects rather than emitting one number, and the scores are
consumed by a non-compensatory rule that decides what is
\emph{admissible} before deciding what is best. Everything left of the
decision rule is standard; the contribution is the box on the right,
and specifically the order in which its cases are tried.

\RAMSD{} replaces one scalar with seven channels, each scored in
$[0,1]$ and each targeting a failure mode a scalar judge cannot
separate (Table~\ref{tab:signal_summary}). Grounding channels follow
claim-verification metrics \citep{min2023factscore} but are computed
online at reward time rather than offline. Channels are
\emph{complementary}, not orthogonal: they target different failures
but are not assumed statistically independent. The point of the
decomposition is auditability ,  for any step, the seven values are a
trace of \emph{why} it scored as it did.

Two channels carry the structural load. \textbf{Logical validity}
$\Rlogic$ asks whether a step is inferentially connected to what
precedes it, independently of whether its content is true ,  a step
can follow correctly from false premises, exactly the Qatar/Dubai
pattern. \textbf{Factual grounding} $\Rfact$ asks the complementary
question, whether the step survives contact with retrieved evidence,
and is the channel Eq.~\eqref{eq:conflation} makes necessary. Among the
supporting five, triangulation requires $M{=}3$ distinct corroborating
sources so one topically-overlapping document cannot certify a claim;
adversarial robustness compares $\Rfact$ against a counterfactually
perturbed step, catching lexical rather than semantic support; and
minimal sufficiency closes the reward-hacking route of padding a step
with unverifiable material.

\begin{table}[t]
\centering
\footnotesize
\setlength{\tabcolsep}{4pt}
\caption{The seven \RAMSD{} channels, each computed from step text,
reasoning context, and retrieved evidence $\mathcal{D}_t$, and each
isolating a judge failure mode a single scalar cannot represent. $c_t$:
token confidence; $e_t$: evidence strength; $n_{\text{indep}}$:
distinct corroborating sources ($M{=}3$); $n_{\text{new}}$: new
unsupported claims ($\beta{=}0.1$).}
\label{tab:signal_summary}
\begin{tabular}{@{}lll@{}}
\toprule
\textbf{Channel} & \textbf{Judge failure mode isolated} &
\textbf{Score} \\
\midrule
1. Logical validity $\Rlogic$ & Chain validity; non-sequiturs &
$(1+f_{\text{NLI}}(s_t \mid Q,s_{<t}))/2$ \\
2. Factual grounding $\Rfact$ & Fluent but unsupported claims &
$(1+\max_{d}f_{\text{NLI}}(s_t\mid d))/2$ \\
3. Epistemic calibration $\Repist$ & Overconfidence; confident hedging &
$1-|c_t-e_t|$ \\
4. Information gain $\Rinfo$ & Verbosity/length bias; stalling &
$(\text{nov}_t+\text{prog}_t)/2$ \\
5. Source triangulation $\Rtriang$ & Single-source corroboration &
$\min(n_{\text{indep}}/M,\,1)$ \\
6. Adversarial robustness $\Radv$ & Surface vs.\ semantic sensitivity &
$\sigma(\Rfact(s_t)-\Rfact(s_t^{\text{counter}}))$ \\
7. Minimal sufficiency $\Rmin$ & Padding with unverifiable claims &
$\exp(-\beta\, n_{\text{new}})$ \\
\bottomrule
\end{tabular}
\end{table}

\subsection{Step-level gating (the variant that failed)}
\label{sec:gate}

Let $\bar R(\sigma_t)=\sum_{i=1}^{7} w_i R_i(\sigma_t)$ with
$\sum_i w_i=1$ and $\mathbf{w}=(.20,.20,.10,.15,.10,.15,.10)$. To block
compensation, logic and factuality act as feasibility constraints with
threshold $\tau=0.5$:
\begin{equation}
\Rstep(\sigma_t)=
\begin{cases}
\bar R(\sigma_t), & \min\{\Rlogic(\sigma_t),\Rfact(\sigma_t)\}\ge\tau,\\
\min\{\Rlogic(\sigma_t),\Rfact(\sigma_t)\}, & \text{otherwise},
\end{cases}
\label{eq:gated_aggregation}
\end{equation}
and the completion reward is
$R_{\mathrm{total}}(y)=3.0\,R_{\mathrm{ORM}}+0.5\,R_{\mathrm{format}}
+\frac1n\sum_t \Rstep(\sigma_t)$, optimized with TRL's
\texttt{GRPOTrainer}; baselines differ only in the reward callback.
Equation~\eqref{eq:gated_aggregation} prevents \emph{step-level}
compensation but still collapses all feasible channels into one number
before the optimizer sees them; empirically it is null
(Section~\ref{sec:negative}), motivating moving the constraint from
the summand to the decision.

\subsection{Evidence-Locked Derive--Gate--Repair}
\label{sec:eldgr}

\ELDGR{} makes evidence an \emph{admissibility} condition at the group
level. For retrieval QA define the strict extractive certificate
$c(a,E)=\ind[a\neq\varnothing \wedge a\ \text{occurs verbatim in}\ E]$
and return
\begin{equation}
a_{\mathrm{ret}} =
\begin{cases}
r, & c(r,E)=1,\ c(m,E)=c(s,E)=0,\\
s, & c(s,E)=1,\ c(m,E)=0,\\
m, & \text{otherwise}.
\end{cases}
\label{eq:eldgr}
\end{equation}
For arithmetic the constraint is executable rather than documentary:
the verifier must expose a derived numeric answer together with a
consistency check, and the branch returns the checked repair $r$ when
it parses and $m$ otherwise. The task-adaptive rule is
\begin{equation}
a_{\ELDGR} =
\begin{cases}
a_{\mathrm{arith}}, & \tau(Q)=\mathrm{arithmetic},\\
a_{\mathrm{ret}}, & \tau(Q)=\mathrm{retrieval},
\end{cases}
\label{eq:task_adaptive}
\end{equation}
with $\tau$ fixed from the benchmark task type, never selected using
test answers.

\paragraph{Why the ordering is the whole method.}
Equation~\eqref{eq:eldgr} is deliberately asymmetric: consensus $m$ is
the default and is displaced only by a \emph{certified} alternative;
the judge's preference $s$ is admitted only when certified \emph{and}
consensus is not; a repair $r$ only when neither $m$ nor $s$ is
certified and $r$ itself is. Auxiliary scores ,  all seven channels
and the judge's scalar ,  rank candidates only inside a feasibility
stratum. This bounds the judge's error profile structurally: to cause a
regression the judge must prefer an answer that is both wrong and
verbatim-present in the evidence while the correct consensus is absent
from it. The certificate is cheap and conservative; it rejects valid
paraphrases, which costs recall, and the arithmetic branch sidesteps
that by checking an executable derivation instead of document
occurrence. Eq.~\eqref{eq:eldgr} also induces a lexicographic group
rank usable for training (certified $\succ$ uncertified; supported
consensus breaks the first tie), but we evaluate it only as a
post-training selector here.

\subsection{What the decomposition does and does not buy}
\label{sec:analysis}

Three component-level results characterize the design choices above.
All describe properties of the \emph{score}; none is a
policy-improvement certificate, which matters because
Section~\ref{sec:negative} reports a null training result.
For the pre-gate weighted score, bilinearity gives
\begin{equation}
\Var(\bar R)=\sum_{i=1}^{7} w_i^2\Var(R_i)
+2\!\!\sum_{1\leq i<j\leq 7}\!\! w_iw_j\Cov(R_i,R_j),
\label{eq:variance}
\end{equation}
so any pair with $w_iw_j>0$ and $\Cov(R_i,R_j)<0$ strictly lowers
$\Var(\bar R)$. The anti-conflation design \emph{manufactures} that
negative covariance: conflated steps in the sense of
Eq.~\eqref{eq:conflation} have high $\Rlogic$ and low $\Rfact$ by
construction, so channel disagreement on exactly the cases of interest
reduces rather than amplifies variance. Two caveats bound the claim.
The deployed gate is nonlinear, so Eq.~\eqref{eq:variance} does
\emph{not} describe $\Rstep$; and the induced score-function bound
$\E\|g_t\|^2\leq G^2\E[\bar R^2]$ (under a $G$-Lipschitz log-policy)
constrains one factor of a second moment, not learning stability.
Two smaller results follow similarly: $\Rfact$ is a monotone transform
of the statistic defining the unsupported-step indicator $U_t$, so
penalizing it penalizes $\E[U_t]$ \emph{conditional on the corpus};
and $\Repist=1-|c_t-e_t|$, an absolute-deviation loss, elicits the
median of the model's evidence-strength belief rather than its full
distribution (Appendix~\ref{app:theory}). Nothing here implies that
optimizing $\Rstep$ improves a policy ,  the analysis characterizes a
random variable, while the empirical failure is about what a
\emph{scalarized} score transmits through GRPO. \ELDGR{} sidesteps that
gap by never scalarizing across feasibility strata.

\section{Experimental Setup}
\label{sec:setup}

\paragraph{Policies and data.}
Base policy \texttt{Qwen2.5-3B-Instruct} with LoRA ($r{=}16$,
$\alpha{=}32$ on \texttt{q,k,v,o\_proj}; $\approx$4.7M trainable
parameters), lr $2\times10^{-5}$, seed 42, 128 new tokens per
completion. GSM8K \citep{cobbe2021training}: train 500, evaluate 500
held out. HotpotQA \citep{yang2018hotpotqa} \texttt{distractor}: train
500, evaluate 300 held out. Retrieval is BM25 over the supporting-facts
corpus ($K{=}5$) for HotpotQA and over 1{,}000 training solutions for
GSM8K, where grounding therefore measures consistency with alternative
valid solution paths rather than external knowledge. NLI throughout is
\texttt{cross-encoder/nli-deberta-v3-small}.

\paragraph{Frozen candidate pool, judge, and statistics.}
One pool per question is built from the four seed-42 full-budget policy
artifacts (GRPO, ORM, PRM, \RAMSD{}), so every selector sees exactly
the same candidates. The judge is a local \texttt{deepseek-r1:7b} via
Ollama with structured JSON output and hidden thinking disabled; it
receives the HotpotQA distractor context and \emph{never} the gold
answer, and homogeneous groups are returned without a judge call. We
report percentile bootstrap intervals over 20{,}000 paired
question-level resamples and two-sided exact McNemar tests; these
quantify evaluation-set uncertainty only, not training-seed variance.
HotpotQA uses official normalization, reporting EM and token F1.

\section{Results}
\label{sec:results}

\subsection{Matched full-budget selection}

\begin{table}[t]
\centering
\footnotesize
\setlength{\tabcolsep}{3.5pt}
\caption{Full-budget matched post-training selection (GSM8K $n{=}500$,
HotpotQA $n{=}300$); all selectors consume the identical frozen
candidate pool. $\Delta$ is \ELDGR{} minus that row, in percentage
points with paired bootstrap 95\% intervals (20{,}000 resamples);
$p$ is the two-sided exact McNemar test on EM. Bold $\Delta$ marks
contrasts resolved at $\alpha{=}0.05$.}
\label{tab:main_results}
\begin{tabular}{@{}lcccccc@{}}
\toprule
& \multicolumn{3}{c}{\textbf{GSM8K}} &
\multicolumn{3}{c}{\textbf{HotpotQA}} \\
\cmidrule(r{0.7em}){2-4}\cmidrule(l){5-7}
\textbf{Selector} & \textbf{Acc.\ (\%)} & \textbf{$\Delta$ [95\% CI]} &
\textbf{$p$} & \textbf{EM / F1} & \textbf{$\Delta$ EM [95\% CI]} &
\textbf{$p$} \\
\midrule
First candidate & $55.4$ & $\mathbf{+2.80}\,[1.20,4.60]$ & $\mathbf{.0026}$ &
$15.33/22.97$ & $+2.00\,[0.33,4.00]$ & $.0703$ \\
Majority vote & $55.8$ & $\mathbf{+2.40}\,[0.40,4.60]$ & $\mathbf{.0357}$ &
$15.33/23.19$ & $\mathbf{+2.00}\,[0.67,3.67]$ & $\mathbf{.0313}$ \\
Scalar judge & $56.8$ & $+1.40\,[-0.20,3.00]$ & $.1435$ &
$15.67/23.49$ & $+1.67\,[0.33,3.33]$ & $.0625$ \\
\midrule
\textbf{\ELDGR{} (ours)} & $\mathbf{58.2}$ & ,  & ,  &
$\mathbf{17.33/25.46}$ & ,  & ,  \\
\bottomrule
\end{tabular}
\end{table}

Two things stand out in Table~\ref{tab:main_results}. First,
\textbf{full judge authority is worth very little}: the scalar judge
beats judge-free majority vote by $1.0$ pp on GSM8K and $0.34$ EM on
HotpotQA, well inside noise. Second, \textbf{the same judge under
\ELDGR{} is worth substantially more}: $+2.4$ pp and $+2.0$ EM over
majority, from identical scores and identical candidates. The
difference is attributable entirely to what the judge may overrule.

On GSM8K the arithmetic derive--repair branch beats first-candidate by
$2.8$ pp with 17 paired wins and 3 losses (exact McNemar
$p{=}0.0026$); the contrast against the scalar judge is positive but
unresolved ($+1.4$ pp, $[-0.2,3.0]$, $p{=}0.143$). This supports
\emph{executable task constraints}, not retrieval-channel importance
,  GSM8K has no external facts to certify. On HotpotQA, evidence
locking gains $2.0$ EM and $2.50$ F1 with seven EM wins and one loss:
the bootstrap interval excludes zero, but the conservative exact test
is borderline ($p{=}0.0703$). F1 gains follow the same ordering.
\ELDGR{} is positive against all three matched selectors on both
datasets, and under-powered only against the stronger judge.

\subsection{Frozen-rule confirmation: a judge that is worse than no judge}
\label{sec:pilot}

The rule in Section~\ref{sec:eldgr} was developed on three disjoint
HotpotQA subsets totalling 120 questions; we froze the prompts,
certificate, and decision order, then evaluated once on a fourth
30-question subset (seed 2029) disjoint from all development data, with
the corrected seed-42 GRPO checkpoint producing the same four
stochastic candidates (temperature $0.7$, top-$p$ $0.9$) for every
selector.

\begin{table}[t]
\centering
\small
\caption{Frozen-rule HotpotQA confirmation split ($n{=}30$), evaluated
once after the decision rule was locked. The unconstrained scalar judge
is the \emph{worst} selector tested ,  below majority vote and below
taking the first sample.}
\label{tab:eldgr_confirmation}
\begin{tabular}{lccc}
\toprule
\textbf{Selector} & \textbf{EM (\%)} & \textbf{F1 (\%)} &
\textbf{$\Delta$ EM vs.\ \ELDGR{} [95\% CI], exact $p$}\\
\midrule
First candidate      & 13.3 & 15.41 & $-13.3\ [-26.7,-3.3]$, $p{=}0.125$\\
Majority vote        & 20.0 & 22.07 & $-6.7\ [-16.7,0.0]$, $p{=}0.500$\\
Scalar judge         & 10.0 & 12.81 & $-16.7\ [-30.0,-3.3]$, $p{=}0.0625$\\
\textbf{\ELDGR{}}    & \textbf{26.7} & \textbf{29.00} & , \\
\bottomrule
\end{tabular}
\end{table}

Table~\ref{tab:eldgr_confirmation} is the result we consider most
relevant to judge reliability. Given full authority over a
four-candidate pool, the scalar judge scores $10.0$ EM ,  $10$ points
below judge-free majority vote and $3.3$ below returning the first
sample. It is not merely uninformative but actively harmful, and
harmful \emph{confidently}, emitting well-formed JSON preferences on
every question. Restricted by Eq.~\eqref{eq:eldgr} to certified
overrides, the same judge yields the best selector in the table ($26.7$
EM, $+16.7$ over its unconstrained self, exact $p{=}0.0625$). Against
majority the $+6.7$ pp difference is unresolved at $n{=}30$ ($p{=}0.5$):
a frozen-rule confirmation that \ELDGR{} improved every tested
selector on this split, not population-level superiority.

\subsection{Decision audit: where the gain comes from}
\label{sec:audit}

On the 30 confirmation questions \ELDGR{} returns the majority answer
on \textbf{22}, the evidence-certified judge choice on \textbf{4}, and
an evidence-certified repair on \textbf{4}, changing \textbf{no}
correct majority decision into an incorrect one and adding two exact
matches. This is the intended shape of a bounded judge: it declines to
act on 73\% of questions, confining the judge's influence to cases
where consensus carries no evidential support. An unconstrained judge
can convert correct consensus into error anywhere; under
Eq.~\eqref{eq:eldgr} only where evidence contains the wrong answer and
not the right one.

\paragraph{Cost.}
Excluding candidate generation, common to all selectors, the scalar and
derive--gate--repair calls average $1.77$s and $2.91$s per pilot
question ($4.68$s total); at full budget the pair averages $1.36$s per
GSM8K and $4.99$s per HotpotQA question, with verifier work skipped
entirely for 352 of 500 homogeneous GSM8K groups (172 of 300 HotpotQA
groups are retrospectively homogeneous and skippable in deployment).
This is an inference-time accuracy--latency trade-off, not a free
lunch.

\section{What Did Not Work}
\label{sec:negative}

\begin{figure}[t]
\centering
\includegraphics[width=0.78\textwidth]{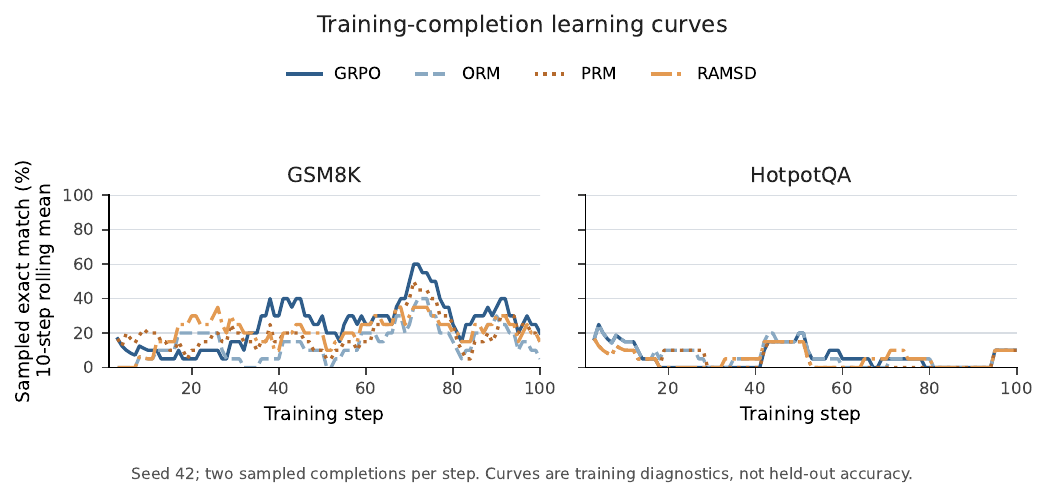}
\caption{Sampled-completion exact match during training (seed 42, two
candidates per step, 10-step rolling mean) ,  on-policy diagnostics,
\emph{not} held-out accuracy. No reward method separates stably, the
basis for treating the gated-reward comparison as null rather than as a
small win or loss.}
\label{fig:learning_curves}
\end{figure}

\paragraph{The gated reward is null.}
Training with Eq.~\eqref{eq:gated_aggregation} did not improve held-out
accuracy over plain GRPO. Figure~\ref{fig:learning_curves} shows why we
decline to read anything into small differences between reward methods:
the on-policy trajectories (two sampled completions per optimizer step,
10-step rolling mean) fluctuate substantially on GSM8K ,  including a
transient GRPO peak near step 72 ,  and are sparse and frequently zero
on HotpotQA. No method separates stably, which means final-checkpoint
selection alone can manufacture or erase a small reported gap. Under
$G{=}2$ the z-normalized advantages are also close to binary, which
plausibly contributes.

\paragraph{The legacy ablation collapses.}
Single-seed GSM8K channel-drop runs fall to $44.6$--$47.4\%$: not just
below full \RAMSD{} ($55.6\%$) but $8$--$11$ pp below GRPO, which uses
none of the channels ($p<3.4\times10^{-6}$ on every row). A collapse of
this shape is incompatible with a marginal-importance reading; zeroing
and renormalizing one weight perturbed the trajectory under a
high-variance $G{=}2$ single-seed protocol. We retain it as a
diagnostic failure, not evidence that each channel matters.

\paragraph{Corrected ablations show no channel is necessary.}
Rerunning every channel drop on HotpotQA with the corrected chat
template, $G{=}4$, and 30 optimizer steps
(Table~\ref{tab:hotpot_ablation}) removes the pathology ,  all rows
land between 18\% and 20\% EM ,  but supplies no support for the
decomposition either. Dropping logic or factuality removes the channel
from both the weighted score and the feasibility gate; other drops zero
and renormalize the corresponding auxiliary weight. No drop differs
from full gated \RAMSD{} by more than one example, every exact paired
McNemar test against full gives $p{=}1.0$, and the factuality,
information, and minimality drops are numerically \emph{above} full. We
therefore cannot claim that any individual channel is load-bearing at
this budget; multi-seed, full-budget ablations remain required.

\begin{table}[t]
\centering
\footnotesize
\caption{Corrected HotpotQA channel-drop pilot (seed 42, $G{=}4$, 30
steps, same 100 evaluation examples for all rows). Deltas are EM
percentage points. Every drop vs.\ full: exact McNemar $p{=}1.0$.}
\label{tab:hotpot_ablation}
\setlength{\tabcolsep}{4pt}
\begin{tabular}{lccc@{\hskip 1.2em}lccc}
\toprule
\textbf{Config} & \textbf{EM/F1} & \textbf{$\Delta$Full} & \textbf{$\Delta$GRPO} &
\textbf{Config} & \textbf{EM/F1} & \textbf{$\Delta$Full} & \textbf{$\Delta$GRPO} \\
\midrule
Full gated \RAMSD{} & 19 / 28.62 & ,  & $+1$ & w/o $\Rinfo$    & 20 / 28.15 & $+1$ & $+2$ \\
GRPO                & 18 / 27.99 & $-1$ & ,  & w/o $\Rtriang$  & 19 / 27.46 & $0$  & $+1$ \\
w/o $\Rlogic$       & 18 / 27.65 & $-1$ & $0$ & w/o $\Radv$     & 18 / 26.53 & $-1$ & $0$  \\
w/o $\Rfact$        & 20 / 29.17 & $+1$ & $+2$& w/o $\Rmin$     & 20 / 28.85 & $+1$ & $+2$ \\
w/o $\Repist$       & 18 / 27.01 & $-1$ & $0$ & & & & \\
\bottomrule
\end{tabular}
\end{table}

The boundary these draw is the point of the paper: the decomposition
earns nothing when summed into a training signal, and something when
used as a \emph{feasibility partition} ,  because a partition cannot
be summed away.

\FloatBarrier
\section{Discussion and Limitations}
\label{sec:discussion}

\paragraph{Bound the judge, don't perfect it.}
Tables~\ref{tab:main_results} and \ref{tab:eldgr_confirmation} say that
judge \emph{accuracy} and judge \emph{usefulness} are only loosely
coupled: the same \texttt{deepseek-r1:7b} scores are worth $-10$ pp or
$+6.7$ pp against majority vote depending only on the rule that
consumes them. Before investing in a better judge, ask what the current
one may overrule ,  a design parameter under direct control. A judge
deployment record should state (i)~the default the judge must beat,
(ii)~the certificate required to beat it, and (iii)~the measured
override rate split by outcome; ours are 22/30 defaults preserved,
extractive occurrence, and 8 overrides with 0 correct-to-incorrect
flips.

\paragraph{Reading the null result correctly.}
Section~\ref{sec:negative} is not evidence that decomposed evaluation
does not help: the decomposition failed where it was summed back into a
scalar and succeeded where it was used to partition. A judge pipeline
that computes seven diagnostics and then averages them has, from the
optimizer's perspective, computed one number.

\paragraph{Limitations.}
\emph{Verifier-relative factuality:} $\Rfact$ and the certificate are
defined against the retrieved corpus and NLI model, so their errors
propagate into the decision. \emph{Certificate scope:} exact-substring
matching rejects supported paraphrases and does not prove the extracted
span satisfies the question's relation. \emph{Statistical power:} one
training seed and one candidate pool; the HotpotQA first-candidate
contrast is borderline ($p{=}0.0703$) and $n{=}30$ cannot resolve the
majority contrast. \emph{Scale:} one 3B policy, one 7B judge, two
benchmarks, with candidate artifacts inheriting a legacy
missing-chat-template defect, so absolute scores are not comparable to
published state of the art. \emph{Responsible use:} evidence-aware
selection is more persuasive and inherits corpus bias; high-stakes
deployments should retain evidence logs and human review.

\FloatBarrier
\section{Conclusion}
\label{sec:conclusion}

On frozen candidate pools, an unconstrained scalar judge is worth
roughly nothing over majority vote at full budget and is the worst
selector we tested on a frozen-rule split; the identical judge under
task-adaptive evidence certificates is the best (58.2\% on GSM8K,
$p{=}0.0026$; 17.33/25.46 EM/F1 on HotpotQA, $p{=}0.0703$) with zero
correct-to-incorrect flips. A judge's error profile is bounded far more
cheaply by constraining what it may overrule than by improving what it
scores. Multi-seed, pre-registered replication is the next step.

\bibliographystyle{plainnat}
\bibliography{references}

@inproceedings{ouyang2022training,
  title={Training language models to follow instructions with
         human feedback},
  author={Ouyang, Long and others},
  booktitle={Advances in Neural Information Processing Systems
             (NeurIPS)},
  volume={35},
  pages={27730--27744},
  publisher={Curran Associates, Inc.},
  address={Red Hook, NY, USA},
  year={2022}
}

@misc{lightman2023let,
  title={Let's verify step by step},
  author={Lightman, Hunter and others},
  year={2023},
  eprint={2305.20050},
  archivePrefix={arXiv},
  note={arXiv:2305.20050}
}

@misc{shao2024deepseekmath,
  title={DeepSeekMath: Pushing the limits of mathematical
         reasoning in open language models},
  author={Shao, Zhihong and others},
  year={2024},
  eprint={2402.03300},
  archivePrefix={arXiv},
  note={arXiv:2402.03300}
}

@misc{cobbe2021training,
  title={Training verifiers to solve math word problems},
  author={Cobbe, Karl and others},
  year={2021},
  eprint={2110.14168},
  archivePrefix={arXiv},
  note={arXiv:2110.14168}
}

@inproceedings{yang2018hotpotqa,
  title={HotpotQA: A dataset for diverse, explainable
         multi-hop question answering},
  author={Yang, Zhilin and others},
  booktitle={Proceedings of the 2018 Conference on Empirical
             Methods in Natural Language Processing},
  address={Brussels, Belgium},
  publisher={Association for Computational Linguistics},
  pages={2369--2380},
  doi={10.18653/v1/D18-1259},
  year={2018}
}

@inproceedings{lewis2020retrieval,
  title={Retrieval-augmented generation for
         knowledge-intensive NLP tasks},
  author={Lewis, Patrick and others},
  booktitle={Advances in Neural Information Processing
             Systems (NeurIPS)},
  volume={33},
  pages={9459--9474},
  publisher={Curran Associates, Inc.},
  address={Red Hook, NY, USA},
  year={2020}
}

@misc{wang2025math,
  title={Math-Shepherd: Verify and Reinforce LLMs
         Step-by-step without Human Annotations},
  author={Wang, Peiyi and others},
  year={2025},
  eprint={2312.08935},
  archivePrefix={arXiv},
  note={arXiv:2312.08935}
}

@inproceedings{min2023factscore,
  title={FActScore: Fine-grained Atomic Evaluation of
         Factual Precision in Long Form Text Generation},
  author={Min, Sewon and Krishna, Kalpesh and Lyu, Xinxi
          and Lewis, Mike and Yih, Wen-tau and Koh, Pang
          and Iyyer, Mohit and Zettlemoyer, Luke and
          Hajishirzi, Hannaneh},
  booktitle={Proceedings of the 2023 Conference on
             Empirical Methods in Natural Language
             Processing},
  address={Singapore},
  publisher={Association for Computational Linguistics},
  pages={12076--12100},
  doi={10.18653/v1/2023.emnlp-main.741},
  year={2023}
}

@misc{asai2024selfrag,
  title={Self-RAG: Learning to Retrieve, Generate, and
         Critique through Self-Reflection},
  author={Asai, Akari and Wu, Zeqiu and Wang, Yizhong
          and Sil, Avi and Hajishirzi, Hannaneh},
  year={2024},
  eprint={2310.11511},
  archivePrefix={arXiv},
  note={arXiv:2310.11511}
}

@misc{nakano2021webgpt,
  title={WebGPT: Browser-assisted question-answering with
         human feedback},
  author={Nakano, Reiichiro and Hilton, Jacob and Balaji,
          Suchir and Wu, Jeff and Ouyang, Long and Kim,
          Christina and Nguyen, Minh and Jiang, Helen and
          Chen, Paul and Xu, Yongchao and others},
  year={2021},
  eprint={2112.09332},
  archivePrefix={arXiv},
  note={arXiv:2112.09332}
}

@misc{yao2023react,
  title={ReAct: Synergizing Reasoning and Acting in
         Language Models},
  author={Yao, Shunyu and Zhao, Jeffrey and Yu, Dian and
          Du, Nan and Shafran, Izhak and Narasimhan,
          Karthik and Cao, Yuan},
  year={2023},
  eprint={2210.03629},
  archivePrefix={arXiv},
  note={ICLR 2023; arXiv:2210.03629}
}

@misc{wang2023selfconsistency,
  title={Self-Consistency Improves Chain of Thought
         Reasoning in Language Models},
  author={Wang, Xuezhi and Wei, Jason and Schuurmans,
          Dale and Le, Quoc and Chi, Ed and Narang, Sharan
          and Chowdhery, Aakanksha and Zhou, Denny},
  year={2023},
  eprint={2203.11171},
  archivePrefix={arXiv},
  note={arXiv:2203.11171}
}

@misc{chern2023factool,
  title={FacTool: Factuality detection in generative AI
         --- a tool augmented framework for multi-task
         and multi-domain scenarios},
  author={Chern, I-Chun and Chern, Steffi and Chen, Shiqi
          and Yuan, Weizhe and Feng, Kehua and Zhou, Chunting
          and He, Junxian and Neubig, Graham and Liu,
          Pengfei},
  eprint={2307.13528},
  archivePrefix={arXiv},
  year={2023},
  note={arXiv:2307.13528}
}

@article{wang2025drm,
  title={From {<Answer>} to {<Think>}: Multidimensional
         Supervision of Reasoning Process for {LLM}
         Optimization},
  author={Wang, Beining and Su, Weihang and Tian, Hongtao
          and Yang, Tao and Zhou, Yujia and Yao, Ting and
          Ai, Qingyao and Liu, Yiqun},
  journal={arXiv preprint arXiv:2510.11457},
  year={2025}
}

@article{wu2025bcrl,
  title={Mitigating {LLM} Hallucination via Behaviorally
         Calibrated Reinforcement Learning},
  author={Wu, Jiayun and others},
  journal={arXiv preprint arXiv:2512.19920},
  year={2025}
}

@article{liu2025gar,
  title={Generative Adversarial Reasoner: Enhancing {LLM}
         Reasoning with Adversarial Reinforcement Learning},
  author={Liu, Qihao and Ye, Luoxin and Ma, Wufei and
          Chou, Yu-Cheng and Yuille, Alan},
  journal={arXiv preprint arXiv:2512.16917},
  year={2025}
}

@inproceedings{azim2025autodspy,
  title={{AutoDSPy}: Automating Modular Prompt Design with
         Reinforcement Learning for Small and Large
         Language Models},
  author={Azim, Nafew and Alam, Abrar Ur and Omar, Hasan Bin
          and Jami, Abdullah Mohammad Muntasir Adnan and
          Ahad, Jawad Ibn and Kabir, Muhammad Rafsan and
          Hossain, Md. Ismail and Rahman, Fuad and Amin,
          Mohammad Ruhul and Rahman, Shafin and Mohammed,
          Nabeel},
  booktitle={Proceedings of the 2025 Conference on Empirical
             Methods in Natural Language Processing:
             Industry Track},
  pages={2881--2896},
  year={2025}
}

@article{lu2025swirl,
  title={{SWIRL}: A Staged Workflow for Interleaved
         Reinforcement Learning in Mobile {GUI} Control},
  author={Lu, Quanfeng and Ma, Zhantao and Zhong, Shuai and
          Wang, Jin and Yu, Dahai and Ng, Michael K. and
          Luo, Ping},
  journal={arXiv preprint arXiv:2508.20018},
  year={2025}
}

@inproceedings{zheng2023judging,
  title     = {Judging {LLM}-as-a-Judge with {MT-Bench} and Chatbot Arena},
  author    = {Zheng, Lianmin and Chiang, Wei-Lin and Sheng, Ying and Zhuang, Siyuan and Wu, Zhanghao and Zhuang, Yonghao and Lin, Zi and Li, Zhuohan and Li, Dacheng and Xing, Eric P. and Zhang, Hao and Gonzalez, Joseph E. and Stoica, Ion},
  booktitle = {Advances in Neural Information Processing Systems (NeurIPS) Datasets and Benchmarks Track},
  year      = {2023},
  note      = {arXiv:2306.05685}
}

@inproceedings{wang2024fair,
  title     = {Large Language Models are not Fair Evaluators},
  author    = {Wang, Peiyi and Li, Lei and Chen, Liang and Cai, Zefan and Zhu, Dawei and Lin, Binghuai and Cao, Yunbo and Liu, Qi and Liu, Tianyu and Sui, Zhifang},
  booktitle = {Proceedings of the 62nd Annual Meeting of the Association for Computational Linguistics (ACL)},
  year      = {2024},
  note      = {arXiv:2305.17926}
}

@inproceedings{panickssery2024selfpref,
  title     = {{LLM} Evaluators Recognize and Favor Their Own Generations},
  author    = {Panickssery, Arjun and Bowman, Samuel R. and Feng, Shi},
  booktitle = {Advances in Neural Information Processing Systems (NeurIPS)},
  year      = {2024},
  note      = {arXiv:2404.13076}
}

@article{gu2024judgesurvey,
  title   = {A Survey on {LLM}-as-a-Judge},
  author  = {Gu, Jiawei and Jiang, Xuhui and Shi, Zhichao and Tan, Hexiang and Zhai, Xuehao and Xu, Chengjin and Li, Wei and Shen, Yinghan and Ma, Shengjie and Liu, Honghao and Wang, Yuanzhuo and Guo, Jian},
  journal = {arXiv preprint arXiv:2411.15594},
  year    = {2024}
}

@misc{zhang2026vacs,
  title   = {{VACS}: Value-Aligned Compositional Shielding for Multi-Agent Reasoning},
  author  = {Zhang, Yiyao and Goel, Diksha and Ahmad, Hussain},
  year    = {2026},
  note    = {Available at SSRN 7115827}
}

@misc{zhang2026meta,
  title   = {Beyond Reactive Agents: Uncertainty-Gated Meta-Reasoning for Tool-Augmented Decision-Making},
  author  = {Zhang, Yiyao and Goel, Diksha and Ahmad, Hussain},
  year    = {2026},
  note    = {Available at SSRN 6997675}
}

@misc{chen2025trader,
  title   = {{3S-Trader}: A Multi-{LLM} Framework for Adaptive Stock Scoring, Strategy, and Selection in Portfolio Optimization},
  author  = {Chen, K. and Ahmad, Hussain and Goel, Diksha and Szabo, Claudia},
  year    = {2025},
  eprint  = {2510.17393},
  archivePrefix = {arXiv},
  primaryClass  = {cs.LG}
}

@misc{santhosh2026healthcare,
  title   = {Comparative Analysis of Large Language Models in Healthcare},
  author  = {Santhosh, S. and Abbas, F. and Ahmad, Hussain and Szabo, Claudia},
  year    = {2026},
  eprint  = {2604.10316},
  archivePrefix = {arXiv},
  primaryClass  = {cs.CL}
}

@article{arifin2026agenticvm,
  title   = {{AgenticVM}: Agentic {AI} for Adaptive Software Vulnerability Management},
  author  = {Arifin, A. and Ahmad, Hussain and Zhang, Yiyao and Goel, Diksha},
  journal = {IEEE Software},
  year    = {2026}
}

\appendix
\section{Positioning Against Recent Multi-Signal Methods}
\label{app:positioning}

\begin{table}[h]
\centering
\footnotesize
\setlength{\tabcolsep}{4pt}
\caption{Positioning against five recent multi-signal or process-feedback
methods. ``Matched here'' means rerun on our frozen candidate pool;
dashes mark systems whose published numbers are not protocol-comparable
and are therefore not reproduced as baselines.}
\label{tab:recent_positioning}
\begin{tabular}{@{}llll c@{}}
\toprule
\textbf{Method} & \textbf{Year} & \textbf{Primary unit} &
\textbf{Main distinction} & \textbf{Matched here} \\
\midrule
DRM \citep{wang2025drm} & 2025 & reasoning dimensions &
confidence/relevance/coherence reward & ,  \\
BCRL \citep{wu2025bcrl} & 2025 & calibrated response &
proper-score confidence and abstention & ,  \\
GAR \citep{liu2025gar} & 2025 & reasoner--discriminator &
dense adversarial process feedback & ,  \\
AutoDSPy \citep{azim2025autodspy} & 2025 & DSP pipeline &
RL-based modular pipeline selection & ,  \\
SWiRL \citep{lu2025swirl} & 2025 & agent trajectory &
staged interleaved multi-agent RL & ,  \\
\midrule
\ELDGR{} (ours) & 2026 & candidate group &
task constraint plus evidence admissibility & yes \\
\bottomrule
\end{tabular}
\end{table}

\section{Design Analysis: Derivations}
\label{app:theory}

\paragraph{Exact variance decomposition.}
For $\bar R=\sum_{i=1}^{7}w_iR_i$ with $\sum_iw_i=1$, $w_i\geq0$, and
$R_i\in[0,1]$,
$\Var(\bar R)=\sum_i w_i^2\Var(R_i)+2\sum_{i<j}w_iw_j\Cov(R_i,R_j)$
by bilinearity. Hence if any pair with $w_iw_j>0$ has
$\Cov(R_i,R_j)<0$ then $\Var(\bar R)<\sum_i w_i^2\Var(R_i)$. Conflated
steps in the sense of Eq.~\eqref{eq:conflation} have high
$\Rlogic$ and low $\Rfact$ by construction, so the anti-conflation
design induces exactly the negative covariance that lowers pre-gate
variance. If additionally the log-policy score is uniformly
$G$-Lipschitz, then $g_t=\phi_t\bar R(\sigma_t)$ satisfies
$\E\|g_t\|^2\leq G^2\,\E[\bar R^2]$, so reducing $\Var(\bar R)$
tightens one factor of the score-function second-moment bound. This is
a bound on an estimator, not a learning-stability guarantee, and it
does not apply to the gated reward $\Rstep$, which is nonlinear.

\paragraph{Unsupported-step rate.}
With $U_t=\ind[\max_{d\in\mathcal{D}_t}f_{\mathrm{NLI}}(s_t\mid
d)\leq\eta]$, the factual-grounding channel is a monotone transform of
the same statistic, $\Rfact=(1+\max_d f_{\mathrm{NLI}})/2$, so
penalizing $\Rfact$ directly penalizes $\E[U_t]$ conditional on the
corpus. The bound is corpus-relative: it says nothing about steps whose
support is absent from $\mathcal{D}_t$.

\paragraph{Calibration channel.}
$\Repist=1-|c_t-e_t|$ is an absolute-deviation loss and therefore
elicits the median of the model's evidence-strength belief rather than
its full predictive distribution. A proper scoring rule would elicit
more, at the cost of a channel that is no longer bounded in $[0,1]$ by
construction.

\section{Reproducibility}
\label{app:repro}
Code, per-question predictions for all four selectors on both datasets,
verifier prompts, timing logs, and the bootstrap/McNemar scripts are
released with the paper. All reported runs use seed 42 for training and
seed 2029 for the confirmation split; the frozen candidate pool is
distributed so that the selector comparison can be rerun without GPU
access.

\end{document}